\documentclass[nonacm,sigconf]{acmart}

\usepackage{graphicx}
\usepackage[straightquotes]{newtxtt}
\usepackage{xspace}
\usepackage{color}
\usepackage{subfig}
\usepackage{hyperref}
\hypersetup{
    colorlinks=true,
    citecolor=black,
    linkcolor=black,
    urlcolor=blue
}
\usepackage{amsthm}
\usepackage{listings}

\newcommand{\mypara}[1]{\par\textbf{#1.}}

\newcommand{\IN}{\textsf{\mbox{\textsc{in}}}\xspace}

\newcommand{\OUT}{\textsf{\mbox{\textsc{out}}}\xspace}
\newcommand{\BLUNDER}{\textsf{\mbox{\textsc{blunder}}}\xspace}
\newcommand{\UNDEC}{\textsf{\mbox{\textsc{undec}}}\xspace}
\newcommand{\OIN}{\textsf{\mbox{\textsc{in{\tt '}}}}\xspace}

\newcommand{\OOUT}{\textsf{\mbox{\textsc{out{\tt '}}}}\xspace}
\newcommand{\OBLUNDER}{\textsf{\mbox{\textsc{blunder{\tt '}}}}\xspace}

\newcommand{\WON}{\textsf{\mbox{\textsc{won}}}\xspace}
\newcommand{\WONPR}{\textsf{\mbox{\textsc{won$_\mathsf{pr}$}}}\xspace}
\newcommand{\WONSC}{\textsf{\mbox{\textsc{won$_\mathsf{sc}$}}}\xspace}
\newcommand{\LOST}{\textsf{\mbox{\textsc{lost}}}\xspace}
\newcommand{\DRAWN}{\textsf{\mbox{\textsc{drawn}}}\xspace}

\newcommand{\figref}[1]{Fig.\,\ref{#1}}
\newcommand{\Figref}[1]{Fig.\,\ref{#1}}
\newcommand{\pos}[1]{\textsf{#1}}
\newcommand{\la}{\ensuremath{\mathbin{\leftarrow}}}
\newcommand{\GR}{\textit{$R_\exists$}\xspace}
\newcommand{\RR}{\textit{$R_\forall$}\xspace}
\newcommand{\pI}{\ensuremath{\mathrm{I}}\xspace}
\newcommand{\pII}{\ensuremath{\mathrm{II}}\xspace}
\newcommand{\becomes}{\ensuremath{\mathbin{:=}}}

\newcommand{\Prov}{\ensuremath{\mathcal P}}
\newcommand{\Pprov}{\ensuremath{\Prov_{\mathsf{pr}}}}
\newcommand{\Aprov}{\ensuremath{\Prov_{\mathsf{ac}}}}

\definecolor{DarkGreen}{rgb}{0,0.45,0}
\definecolor{DarkRed}{rgb}{0.8,0,0}
\definecolor{DarkYellow}{rgb}{0.6,0.6,0}
\definecolor{DarkGray}{rgb}{0.6,0.6,0.6}

\theoremstyle{definition}  
\newtheorem{definition}{Definition}

\copyrightyear{2025}
\acmYear{2025}
\setcopyright{cc}
\setcctype{by-nd}
\acmConference[PW'25]{ProvenanceWeek}{June 22--27, 2025}{Berlin, Germany}
\acmBooktitle{ProvenanceWeek (PW'25), June 22--27, 2025, Berlin,
Germany}\acmDOI{10.1145/3736229.3736270}
\acmISBN{979-8-4007-1941-7/2025/06}

\author{Bertram Lud\"{a}scher}
\email{ludaesch@illinois.edu}
\affiliation{%
  \institution{University of Illinois}
  \city{Urbana-Champaign}
  \state{IL}
  \country{USA}
}

\author{Yilin Xia}
\email{yilinx2@illinois.edu}
\affiliation{%
  \institution{University of Illinois}
  \city{Urbana-Champaign}
  \state{IL}
  \country{USA}
}

\author{Shawn Bowers}
\email{bowers@gonzaga.edu}
\affiliation{%
  \institution{Gonzaga University}
  \city{Spokane}
  \state{WA}
  \country{USA}
}

\begin{abstract}
The rule $\pos{Defeated}(x) \la \pos{Attacks}(y,x),\, \neg \, \pos{Defeated}(y)$, evaluated under the  well-founded semantics (WFS), yields a unique 3-valued (skeptical) solution of an abstract argumentation framework (AF): 
An argument $x$ is \emph{defeated} (\OUT) if there exists an undefeated argument $y$ that attacks it. For 2-valued (stable) solutions, this is the case iff $y$ is \emph{accepted} (\IN), i.e., if \emph{all} of $y$'s attackers are defeated. Under WFS, arguments that are neither accepted nor defeated are \emph{undecided} (\UNDEC). 
As shown in prior work, well-founded  solutions (a.k.a.\ \emph{grounded labelings}) ``explain themselves'': The \emph{provenance} of arguments is given by subgraphs (definable via regular path queries) rooted at the node of interest. This provenance is closely related to \emph{winning strategies} of a two-player argumentation game. 

We present a novel approach for extending this provenance to stable AF solutions. Unlike grounded solutions, which can be constructed via a bottom-up \emph{alternating fixpoint procedure}, stable models often involve (non-deterministic) \emph{choice} as part of the search for models. Thus, the provenance of stable  solutions is of a different nature, and reflects a  more expressive \emph{generate \& test} 
paradigm. 
Our  approach identifies minimal sets of \emph{critical attacks},  pinpointing choices and assumptions made by a stable model. These critical attack edges provide additional  insights into the provenance of an argument's status, combining well-founded derivation steps with choice steps. Our approach can be understood as a form of diagnosis that finds  minimal  ``repairs'' to an  AF graph such that the well-founded solution of the repaired graph coincides with the desired stable model of the original AF graph.
\end{abstract}


\title{Choices and their Provenance: Explaining Stable Solutions of Abstract Argumentation Frameworks}

\begin{document}

\maketitle

\section{Introduction}

An argumentation framework (AF) \cite{dung1995acceptability} is a directed graph whose nodes and edges represent abstract \emph{arguments} and a binary \emph{attack relation}, respectively.
 AFs offer well-established, formal 
 approaches for representing and reasoning about conflict, having applications, e.g., in case law~\cite{BenchCapon20}, medical decision making~\cite{Longo2013}, computational linguistics \cite{Cabrio2012}, and multi-agent systems \cite{Rahwan2009}. 
An AF models the conflicts between arguments via the given attack relation. One purpose of computational argumentation is to find \emph{solutions}, i.e., sets of arguments that are \emph{conflict-free} (do not attack each other) and that defend each other against outside attacks. More precisely, an argument $x$ is \emph{accepted} (under a given AF semantics) if each of its attackers $y$ is \emph{defeated} (i.e., counter-attacked by an accepted argument), and \emph{defeated} if it is attacked by at least one accepted argument. 
Two prominent AF semantics are the skeptical, 3-valued \emph{grounded semantics}, and the \emph{credulous} (or \emph{brave}) \emph{stable semantics}. The grounded semantics provides well-founded derivations for the acceptance (and defeat) of arguments that can be computed from the bottom-up. 
The stable semantics can additionally \emph{assume} or \emph{choose} arguments to be accepted, provided the resulting model satisfies certain formal conditions (see below). 

Consider the simple AF example in \Figref{fig:input-af}: Argument \pos A attacks \pos B, which attacks \pos C; and arguments \pos C and \pos D attack each other.
\begin{figure}[!h]
  \centering
  \subfloat[Input AF]{
    \includegraphics[scale=.35]{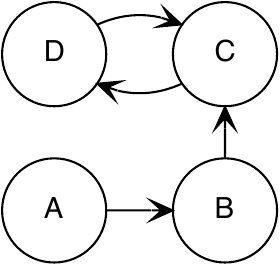}
    \label{fig:input-af}
  }
\hfill
  \subfloat[Grounded $S_0$]{
    \includegraphics[scale=.35]{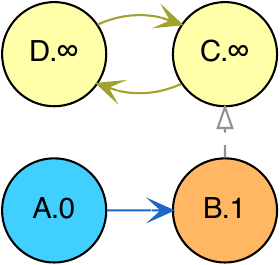}
    \label{fig:simple-wf}
  }
\hfill
  \subfloat[Stable $S_1$]{
    \includegraphics[scale=.35]{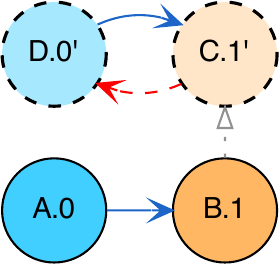}
    \label{fig:simple-s1}
  }
\hfill
  \subfloat[Stable $S_2$]{
    \includegraphics[scale=.35]{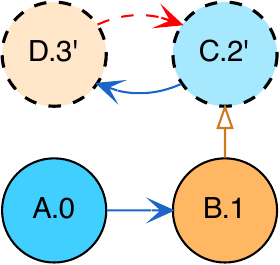}
    \label{fig:simple-s2}
  }
  \caption{\small AF solutions: (a) Input AF, (b) 3-valued grounded solution $S_0$; (c), (d): stable solutions $S_1$ and $S_2$ (as \emph{provenance overlays}).}
  \label{fig:simple}
\end{figure}

\noindent The \emph{grounded solution} is shown in \Figref{fig:simple-wf}: Since \pos A is a source node, it is not attacked by any other argument  and thus can be accepted immediately and labeled \IN (blue color). Because \pos A is \IN and attacks \pos B, the latter is defeated and labeled \OUT (orange). We also label decided nodes with a natural number, their \emph{length}. Here \pos {A.0} and \pos {B.1} indicate well-founded derivations with 0 and 1 inference steps, respectively. What is the acceptance status of \pos C and \pos D? Since \pos B is defeated, the attack $\pos B {\to} \pos C$ fails and can be ignored. This leaves \pos C and \pos D, which attack each other. There is no well-founded derivation that establishes the defeat or acceptance of either \pos C or \pos D, and  the grounded labeling is \emph{undecided} for both (yellow). 
The label ``$\infty$'' indicates a potentially infinite dialogue: there is no finite path of arguments that conclusively resolves the status of  \pos C and \pos D.\footnote{Instead, both parties can force an infinite argumentation game, i.e., a \emph{draw}, see \cite{BowersXiaLudaescherSAFA24}.}

Unlike classical logic, which aims at a single ground truth, formal argumentation explores sets of arguments that stand together---mutually supporting and defensible in light of attacks \cite{dung1995acceptability,BenchCaponModgil09}. The goal is not merely to determine what is definitively true, but to explore which positions a rational agent could justifiably hold, given the attacks and defenses in an AF. Therefore multiple, alternative views (\emph{solutions}) may exist, as exhibited by the two {stable models} in  \figref{fig:simple-s1} and \ref{fig:simple-s2}. Note that what is definitely \IN or \OUT in the grounded solution (\pos A and \pos B, respectively) remains so in all stable solutions. However, what was \UNDEC in the former (\pos C and \pos D) is now decided (\IN or \OUT) in the latter: \figref{fig:simple-s1} is a stable model in which \pos D is \IN and  \pos C is \OUT. Alternatively, in \figref{fig:simple-s2}, \pos C is \IN and \pos D is \OUT. 

\mypara{Choices} How can we explain these alternative solutions? What is the provenance that justifies these alternate views? The crux of the stable  semantics \cite{gelfond_stable_1988} (and the source of its expressive power) is that it can  \emph{guess} a solution and then \emph{check} whether all constraints are satisfied (eliminating incorrect guesses along the way).  \figref{fig:simple-s1} shows that choosing $\IN(\pos D)$ and $\OUT(\pos C)$  provides a stable solution: \pos C is \OUT due to a successful attack from \pos D, which is \IN because the attack from \pos C fails. In \figref{fig:simple-s2}, using a very similar line of reasoning, the \emph{opposite view} is presented, i.e., \pos D is \OUT because its attacker \pos C is \IN, which in turn is justified by not having any attacker that is \IN. Thus, \emph{both} stable solutions can be justified, albeit by making different assumptions (and without additional evidence). 

\mypara{Justifying Choices} We can justify these choices is by finding minimal sets of edges that---\emph{if} ignored or removed---\emph{would} lead to the desired solution, but now in a {well-founded} way. In \figref{fig:simple-s1} and \ref{fig:simple-s2} these \emph{critical attack edges} (depicted in red) are $\Delta_{1} = \{\pos C{\to} \pos D\}$ and  $\Delta_{2} = \{\pos D{\to} \pos C\}$, respectively. We can justify stable solution $S_1$ (or $S_2$) by assuming the attacks in $\Delta_{1}$  (or $\Delta_{2}$) are ``less important'' in the given AF graph, thereby selecting a  particular solution instead of another one. Put another way: if the critical attack edges in  $\Delta_{i}$ were \emph{suspended} (ignored or eliminated), the grounded solution $S'_i$ would match the desired stable solution $S_i$ and the debate about which solution is ``the right one'' would be over. In general, a given stable solution $S_i$ may have multiple, alternative sets of critical attacks $\boldsymbol{\Delta}_i = \{\Delta_{i,1}, \dots, \Delta_{i,n}\}$. Selecting a specific $\Delta_{i,j}$ for $S_i$ allows users to diagnose an AF by finding nodes and edges that may play a fundamental role in the chosen stable solution $S_i$.

\begin{figure*}[!t]
  \centering
  \subfloat[Layered \emph{skeptical} (grounded) solution $S_0$]{
    \includegraphics[width=.31\textwidth]{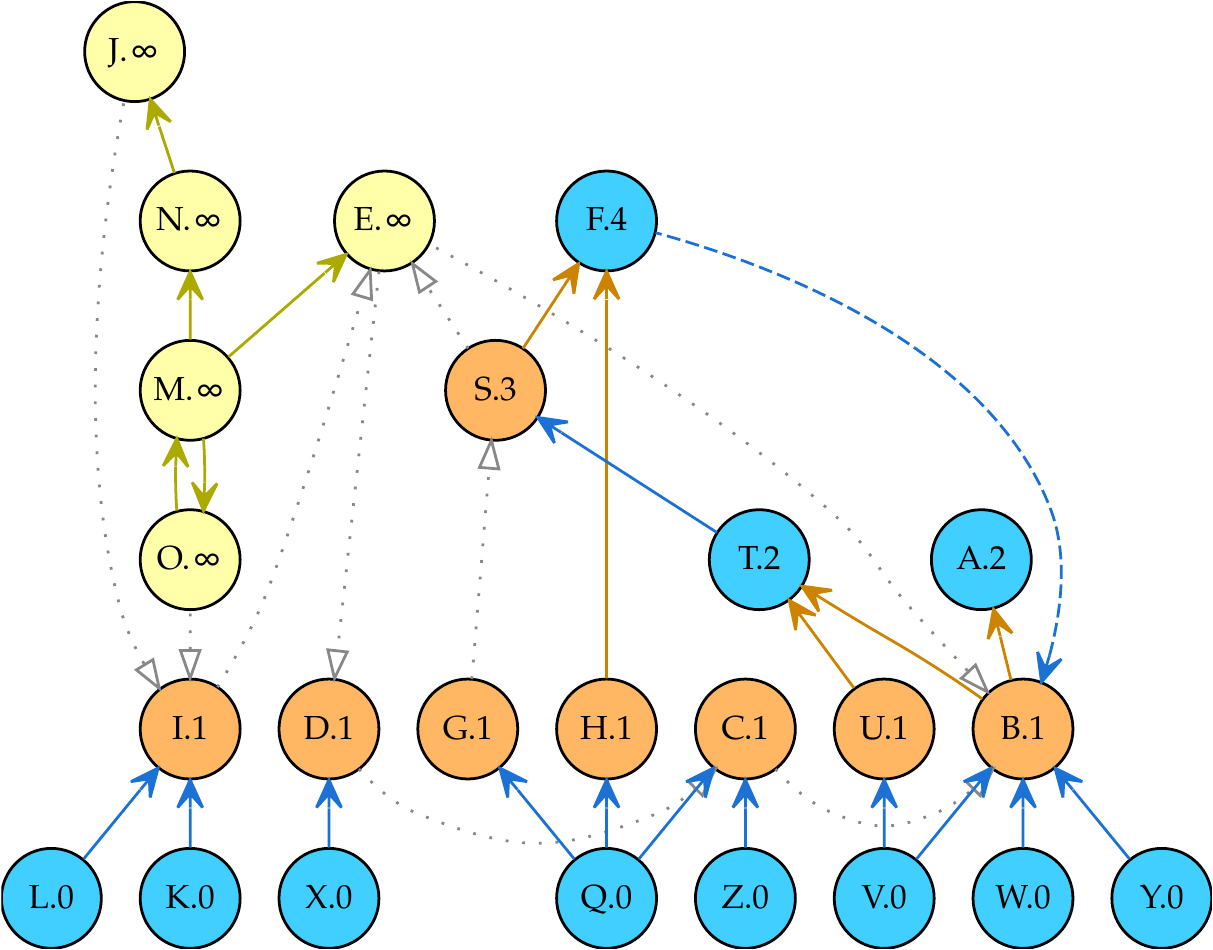}
    \label{fig:layered}}
\hspace{.01\textwidth}
  \subfloat[Overlay: \emph{credulous} (stable) resolution $S'_{1,1}$]{
    \includegraphics[width=.31\textwidth]{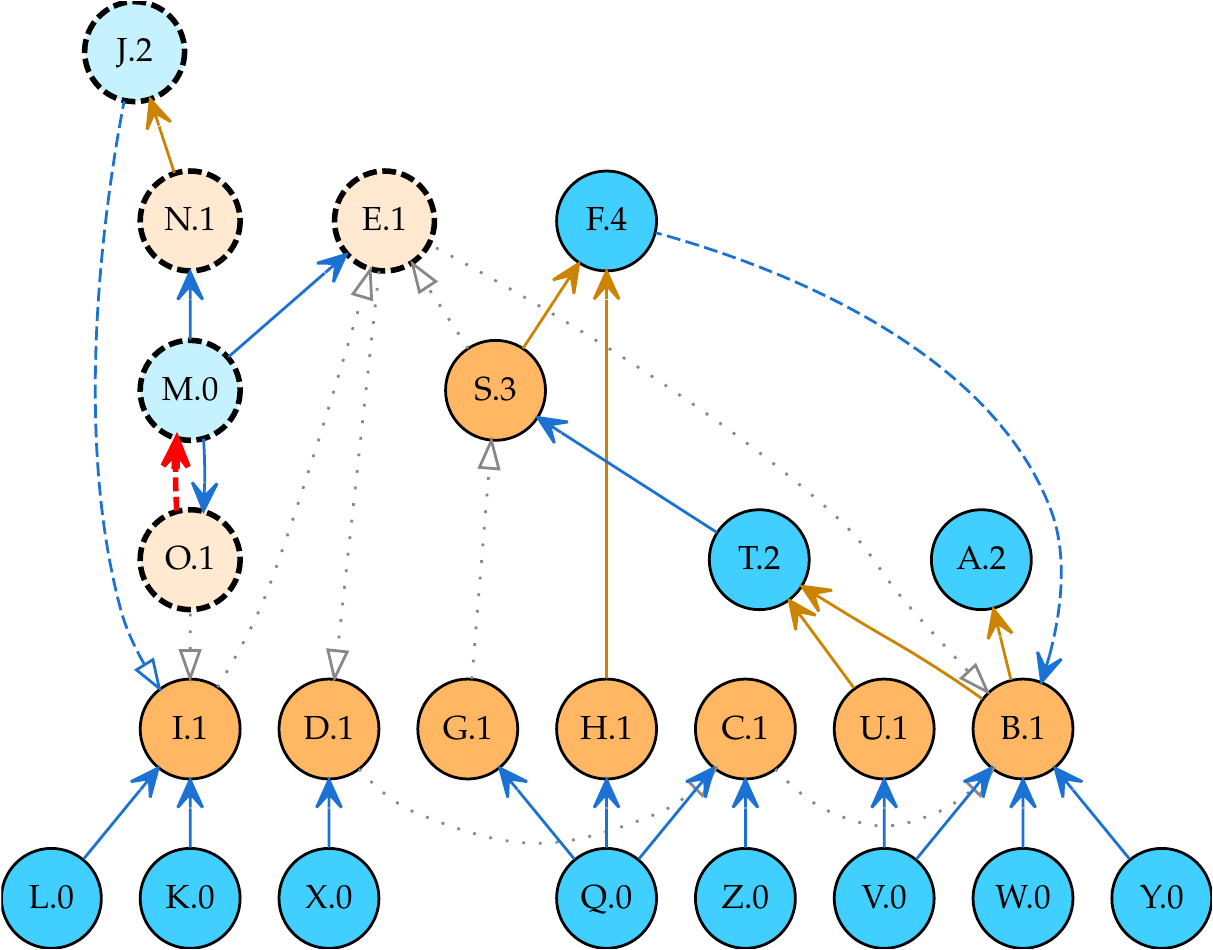}
    \label{fig:stable-1}}
  \hspace{.01\textwidth}
  \subfloat[Overlay: \emph{credulous} (stable) solution $S'_{2,1}$]{
    \includegraphics[width=.31\textwidth]{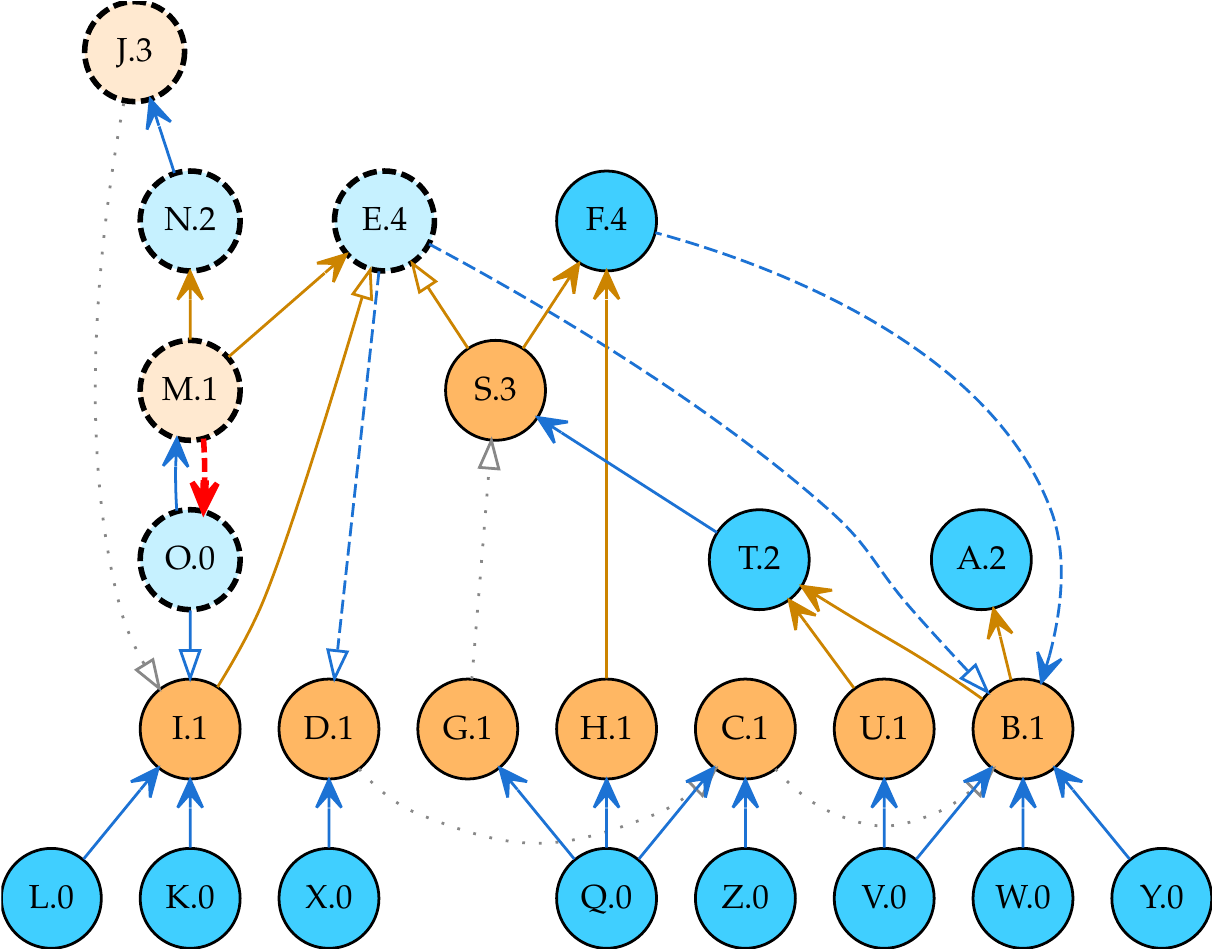}
    \label{fig:stable-2}}
 \caption{AF solutions for \emph{Wild Animals} cases \cite{bench-capon_representation_2002}: (a) The ranked layout of $S_0$ uses the \emph{length} of arguments: e.g., \textsf{F.4} requires at most four rounds of argumentation to prove that \pos F is accepted (\IN). Yellow nodes (length\,=\,$\infty$) are undecided (\UNDEC). 
  Distinct \emph{edge types} are used to account for their semantic roles~\cite{BowersXiaLudaescherSAFA24}.
 The  \emph{overlays} in (b) and (c) represent {alternative} {resolutions} $S'_{1,1}$ and $S'_{2,1}$: The \UNDEC nodes \pos E, \pos J, \pos M, \pos N, \pos O  in (a) have been \emph{decided} (\pos M is \IN, \pos O is \OUT in $S_1$; vice versa in $S_2$). These choices are {explained} by sets of \emph{critical attacks} (red edges) $\boldsymbol{\Delta}_1=\{\Delta_{1,1}{:}\,\{\pos O\to \pos M \}\}$ and $\boldsymbol{\Delta}_2=\{\Delta_{2,1}{:}\,\{\pos M\to \pos{O} \}\}$, i.e., minimal sets of (temporarily) \emph{suspended} edges: when suspensions are applied,  2-valued  grounded solutions $S'_{1,1}$, $S'_{2,1}$ are obtained for $S_1$ and $S_2$.}
 \label{fig:3panel}
\end{figure*}

\mypara{Provenance Overlays} Consider again  \Figref{fig:simple}, its (skeptical) grounded solution $S_0$, and the two stable solutions $S_1$ and $S_2$. In \cite{BowersXiaLudaescherTAPP24} we analyzed well-founded \emph{game provenance} and its structure, and showed how it can be obtained using \emph{regular path queries} (RPQs) over solution labelings. In  \cite{BowersXiaLudaescherSAFA24}, we showed that this rich well-founded provenance  carries over to grounded labelings of abstract AFs. Here, we extend this prior work \cite{BowersXiaLudaescherSAFA24,BowersXiaLudaescherTAPP24,xia_layered_vis_demo_2024} to \emph{stable extensions} of AFs by developing methods to compute and visualize new forms of provenance for stable AF solutions. While the traditional way to describe stable solutions is to simply list the \IN (= accepted) arguments, i.e., $S_1 = \{\pos A,\pos D\}$ and $S_2 = \{\pos A,\pos C\}$, our approach produces more informative hybrid \emph{provenance overlays}: The stable solutions in \figref{fig:simple-s1} and \ref{fig:simple-s2} depict a combination of well-founded provenance information (i.e., different node and edge labels through color-coding and length-labels) and new \emph{choice provenance} as minimal sets $\Delta_{1,1}$ and $\Delta_{2,1}$ of critical attacks. Our  approach therefore ``looks deeply'' into the structure of AFs to help explain the ambiguous acceptance status of arguments.

\subsection{Motivating Example and Overview} 

\Figref{fig:layered} is an example AF for the \emph{Wild Animals} cases of \cite{bench-capon_representation_2002}. It models legal arguments concerning hunting rights and property ownership across various historic legal cases. Each argument (\pos A, \pos B, $\dots$) is colored according to its solution status under the grounded semantics: \IN-labeled (accepted) arguments are blue, \OUT-labeled (defeated) arguments are orange, and \UNDEC-labeled (undecided) arguments are yellow. The layout of \figref{fig:layered} uses a \emph{layered visualization} based on the \emph{length} of nodes. The bottom layer consists of the nodes that are trivially labeled \IN because they have no attackers (shown with length 0), the next layer consists of \OUT nodes (length 1) that are defeated due to a length 0 attacker, etc. \UNDEC arguments (length ``$\infty$'') are displayed outside of the layering. The arguments that determine (and thus \emph{explain}) an \IN or \OUT labeled argument $x$ are located at layers \emph{below} $x$. For instance, while \pos F in \Figref{fig:layered} attacks \pos B, \pos B's defeat is known before \pos F's label is determined due to \pos V, \pos W, and \pos Y. The ranked layout makes explicit the well-founded (and thus ``self-explanatory'') derivation structure of the grounded semantics. 

Attack \emph{types} are displayed according to their role in determining argument labels \cite{BowersXiaLudaescherSAFA24} (see \Figref{fig:attack_types} for details): Successful (blue) attacks are  either \emph{primary} (solid blue) or \emph{secondary} (dashed blue). Secondary attacks point to arguments with smaller lengths: e.g., \pos F's attack on \pos B, whose defeat was already established in a lower layer. In contrast, dotted gray edges (\emph{blunders} in the associated game) are irrelevant for the acceptance status of a node $x$, i.e., they are not part of $x$'s provenance. A minimal explanation of  argument $x$ ignores secondary and blunder attacks, so these edges are de-emphasized in the layered visualization.  

To disambiguate the \UNDEC portion of a grounded solution, all {stable} solutions ($S_1$ and $S_2$) can be enumerated, see \Figref{fig:stable-1} and \ref{fig:stable-2}. 
Each stable solution represents a choice for resolving the circular (direct or indirect) conflicts that lead to \UNDEC arguments in the grounded semantics. 
Each solution $S_i$ can be visualized and explained as an \emph{overlay} $S'_{i,j}$, i.e., $S_i$ may have multiple minimal attack sets $\Delta_{i,j}$: The {overlays}  in \Figref{fig:stable-1}\,{\&}\,\ref{fig:stable-2} use the same layered visualization, but \UNDEC arguments are colored according to their acceptance status in the chosen solution $S_i$ (with lighter colors and dashed outlines). The mutual attack between $\pos M$ (\emph{mere pursuit is not enough}) and $\pos O$ (\emph{bodily seizure is not necessary}) defines the  critical attacks $\pos O \to M$ and $\pos M \to \pos O$, respectively, which are highlighted in red in \Figref{fig:stable-1}\,\&\,\ref{fig:stable-2}. In this case, $\mathbf{\Delta_1} = \{\Delta_{1,1}\}$ and $\mathbf{\Delta_2} = \{\Delta_{2,1}\}$ with $\Delta_{1,1} = \{\pos O{\to}\pos M\}$ and $\Delta_{2,1} = \{\pos M{\to}\pos O\}$, which give rise to the overlays $S'_{1,1}$ and $S'_{2,1}$, respectively. The length and edge type changes of the resulting grounded solution (with critical attacks suspended) are also shown in the corresponding overlays.  

Critical attacks facilitate new use cases for AF reasoning that complement earlier Value-based and Extended-AF approaches \cite{BenchCaponModgil09}. Whereas the latter assume that users already know which edges to attack, our approach systematically generates all such target edges, thus providing a deeper semantic analysis of AF ambiguity than any state-of-the art system we are aware of. 

\subsection{Paper Organization} 
The rest of this paper is organized as follows. Section~\ref{sect:argumentation-frameworks} provides basic definitions of AFs and their semantics. Section~\ref{sect:game-prov} describes our prior work on provenance related to two-player combinatorial games and AFs.  Section~\ref{sect:critical} introduces the notion of critical attack edges and their use in constructing provenance overlays. 
We provide additional examples in Section~\ref{sect:critical} and describe approaches for computing minimal critical attack sets for stable solutions, including an answer set programming (ASP) implementation in Clingo. Section~\ref{sect:discussion} concludes with a discussion of related work on AF explanation techniques as well as our plans for future work. We note that the approaches described in this paper are being used to extend the PyArg framework \cite{odekerken2023pyarg} to provide new explanation and visualization techniques for AFs \cite{xia2025afxray}.  

\section{Abstract Argumentation Frameworks}
\label{sect:argumentation-frameworks}

This section briefly recalls basic definitions of abstract argumentation
frameworks \cite{dung1995acceptability,baroni_handbook_2018}.

\begin{definition}
  An \emph{argumentation framework} (AF) is a finite digraph
  $G=(V,E)$ where the nodes $V$ represent  \emph{arguments} and edges
  $(x,y)\in E \subseteq V{\times}V$ (denoted $x {\to} y$) represent
  \emph{attacks}.
\end{definition}

Let $G = (V,E)$ be an AF. The main reasoning task given $G$ is to determine a set $S \subseteq V$ of arguments to accept. A basic property of such a set $S$ is that it is \emph{conflict free}. 

\begin{definition}
  $S \subseteq V$ is \emph{conflict free} if no two arguments in $S$ attack each
  other. 
\end{definition}

Another basic property of $S$ is that it is \emph{admissible}. A set $S$ is said to \emph{attack} an argument $y \in V$ if $y$ is attacked by at least one argument $x\in S$. Similarly, $S$ is said to \emph{defend} an $x \in V$ if $S$ attacks all of $x$'s attackers. The arguments defended by $S$ are represented by the \emph{characteristic function}, which is used to define admissibility.

\begin{definition} 
For $S \subseteq V$, the \emph{characteristic function} $F : 2^V \to 2^V$ is $F(S) = \{x \mid S \text{ defends } x\}$.
\end{definition}

\begin{definition}
$S \subseteq V$ is \emph{admissible} if it is
  both conflict free and $S \subseteq F(S)$, i.e., if $S$ defends at least itself.
\end{definition}

An AF can have many admissible sets, referred to as \emph{extensions}. Different classes of extensions give rise to different
\emph{extension semantics}. An important, albeit large class are the \emph{complete extensions}. 

\begin{definition}
$S \subseteq V$ is a \emph{complete extension} if it is conflict free and $S = F(S)$, i.e., $S$ exactly defends itself. 
\end{definition}

A given AF can have many complete extensions, which include both the grounded and stable extensions.

\begin{definition}
    $S \subseteq V$ is the \emph{grounded extension} if it is complete and subset minimal. 
\end{definition}

\begin{definition}
    $S \subseteq V$ is a \emph{stable extension} if it is complete and it attacks every $x \in V \setminus S$. 
\end{definition}

In the following, we consider alternative definitions of the grounded and stable extensions via logic-based semantics as well as the notion of argument labelings. 

\subsection{Grounded and Stable Extensions}
\label{sect:grounded}

Dung \cite{dung1995acceptability} showed that the well-founded model of the following Datalog program exactly gives an AF's grounded extension.
\begin{equation}
\begin{array}{@{}r@{~}c@{~}l}
  \pos{Defeated}(x) & \la & \pos{Attacks}(y, x),\, \pos{Accepted}(y). \smallskip \\
  \pos{Accepted}(y) & \la  & \neg \, \pos{Defeated}(y). 
\end{array}
\tag{$P_{\pos{AF2}}$}
\end{equation}
Given an AF $G=(V,E)$, the first rule states that an argument $x$ is defeated if there
is an accepted argument $y$ that {attacks} $x$. The second rule specifies that an argument is accepted if it is not defeated.  The subgoal $\pos{Accepted}(y)$ in the first rule can also be replaced with
the negated atom from the second rule, resulting in an  equivalent single-rule program: 
  \begin{equation}
  \pos{Defeated}(x) \la \pos{Attacks}(y, x),\, \neg \, \pos{Defeated}(y). 
  \tag{$P_\pos{AF1}$}
\end{equation}

\noindent
Similarly, Dung \cite{dung1995acceptability} also showed that $P_\pos{AF1}$ (equivalently $P_\pos{AF2})$ when evaluated under the stable model semantics yields exactly the stable extensions.  

\subsection{Solution Labelings}

Given an AF, we can consider its extensions as its ``solutions'' (under the given extension semantics). An alternative approach is to \emph{label} the arguments according to their \emph{acceptance status} \cite{caminada2006issue,caminada_strong_2020}:

\begin{definition}
Let $G = (V,E)$ be an AF. A \emph{labeling} of $G$ is a function
  $\mathit{Lab} : V \to \{\IN, \OUT, \UNDEC\}$ that
  assigns a label of \IN, \OUT, or \UNDEC to arguments. 
Label values correspond to whether an argument is \IN an extension
(\emph{accepted}), \OUT of an extension (\emph{defeated}), or
neither in nor out (\UNDEC).
\end{definition}

\section{Provenance of Games and Grounded AFs}
\label{sect:game-prov}

Provenance information helps explain the status of an argument in the AF's solution under a given semantics. These explanations are closely tied to dialogues in an AF, i.e., a back-and-forth of argument attacks made by an opponent and proponent. Dialogues (also referred to as discussions) are also closely tied to two-player combinatorial games \cite{BowersXiaLudaescherSAFA24,dung1995acceptability,caminada2018argumentation_as_discussion}. Here we briefly describe our prior work on provenance for ``classic'' combinatorial games, and how this work can be directly applied to AFs using a straightforward correspondence between games and dialogues. 

\subsection{Win-Move Games}

The rule $P_\pos{AF1}$ from Sect.~\ref{sect:argumentation-frameworks} can be equivalently rewritten in a ``reversed edges'' form:
\begin{equation}
  \pos{Defeated}(x) \la \pos{AttackedBy}(x,y),\, \neg \, \pos{Defeated}(y). 
  \tag{$P_\pos{AF'}$}
\end{equation}
An edge $x{\to}y$ in the graph then means that argument $x$ is \emph{attacked-by} an argument $y$. The resulting rule ($P_\pos{AF'}$) is equivalent to the single-rule ``win-move'' (WM) program  from \cite{van1991well}: 
\begin{equation}
  \pos{Win}(x) \la \pos{Move}(x, y), \, \neg \, \pos{Win}(y). \tag{$P_\pos{WM}$}
\end{equation}
Under the well-founded semantics, $P_\pos{WM}$ solves standard two-player combinatorial games. In their general form, these games are played by the two players Player~I and Player~II that, after agreeing on a starting position (node), alternate turns moving a pebble from one position to another along the edges (moves) of a board (represented by a digraph). Player~I makes the first move. Play ends when the player without a move loses, which means their opponent wins. 
A particular \emph{play} $\pi$ of a game is an alternating sequence of moves by the players: 
\begin{displaymath}
x_0 \stackrel{\pI}{\to} x_1 \stackrel{\pII}{\to} x_2
\stackrel{\pI}{\to} \cdots
\end{displaymath}

\noindent
The \emph{play length} $|\pi|$ is the number of moves in a play. 

\begin{definition}
 A play $\pi$ is
\emph{complete} if either $|\pi|=\infty$ (repeating moves\footnote{On
  finite graphs, cycles are necessary (but not sufficient) for a play to
  end in a draw.}) or $\pi$ ends after $|\pi|=n$ moves in a terminal
node. A player who cannot move \emph{loses} (so-called \emph{normal
  play}) and the opponent \emph{wins}. If $|\pi|=\infty$ the play is a
\emph{draw}.
\end{definition}

In win-move games, the outcome of a game is based on the starting position and the moves the two players can make. 

\begin{definition}
The \emph{value} of a position $x$ is
  \WON if a player can force a win from $x$, independent of
  the opponent's moves; $x$'s value is \LOST if there are no moves to
  play or if the opponent can force a win; and $x$ is \DRAWN if neither player can force a win.
\end{definition} 

Intuitively, the \emph{value} of a position (\WON,
\LOST, or \DRAWN) only depends on ``best moves''
(optimal play). In particular, ``bad moves'' (\emph{blunders}), which weakens the player's outcome, do not affect the position value. A player that can force a win from a position has a \emph{winning
  strategy}: a set of moves they can make that lead to a loss for
their opponent regardless of the moves their opponent makes. Determining the value of each position results in a \emph{solved
game}, which is represented via a total labeling function $\lambda$.
\begin{definition}
  The \emph{solution} labeling
  $\lambda : V \to \{\WON, \LOST, \DRAWN\}$
   of game $G = (V,E)$ yields the position values. A \emph{solved game} is denoted $G = (V,E,\lambda)$. 
\end{definition}
It is well known that games can be solved by
iterating the following two rules\footnote{Initially,
  $\lambda(x) \becomes \emptyset$ for each position $x \in V$.}.
\begin{itemize}
\item $\lambda(x) \becomes \LOST$ ~ if $\forall y:  (x,y)  \in E$ implies
  $\lambda(y) = \WON$. \hfill (\RR)
\item $\lambda(x) \becomes \WON$ ~ if $\exists y: (x,y) \in E$ and
  $\lambda(y) = \LOST$. \hfill (\GR)
\end{itemize}
The \emph{forall rule} (\RR) states that a position $x$ is \LOST
if \emph{all} of $x$'s followers $y$ have
already been \WON: No matter which follower
$y$ of $x$ a player moves to, the opponent can force a win from
$y$. The \emph{exists rule} (\GR) states that a position $x$ is \WON if \emph{at least one} of $x$'s followers $y$ has
already been \LOST: A player can thus
choose to move from $x$ to such a $y$, leaving the opponent in a lost
position. 

On an unlabeled graph $G$, the first applicable rule is \RR:
Terminal positions have no moves and so \RR's condition is
vacuously true. The result is that each such terminal position is
assigned \LOST. In the next iteration, \GR becomes applicable,
assigning \WON to all positions that have have at least one
direct move according to the \GR condition to a terminal
position.  The second iteration of \RR labels positions whose moves
all lead to positions previously won. The second iteration of \GR then
assigns labels to positions with at least one move to a lost
position. This stage-wise iteration eventually converges to a
\emph{fixpoint}, and any remaining nodes are drawn positions
\cite{kohler_first-order_2013,ludascher2023games}. The underlying
process of iterating \RR and \GR is equivalent to the classic
\emph{backward induction} procedural approach for solving games. Similarly, solving a game by evaluating the rule $P_\pos{WM}$ under
the \emph{well-founded semantics}~\cite{van1991well}
can be implemented via the \emph{alternating fixpoint procedure}
(AFP)~\cite{van1993alternating}.

\subsection{Explaining Values with Provenance} 

{\em Provenance}  $\Prov(x)$ represents an explanation of why and
how a position $x$ has a particular value in a (solved) game. To
compute $\Prov(x)$, we first add additional provenance information to solved games, including (optimal) position lengths and provenance-specific edge labels. We
then use the provenance information to construct the explanation
$\Prov(x)$ as a subgraph of the solved game rooted at $x$. 

\begin{definition}
  Let $G = (V,E,\lambda)$ be a solved game.
  The \emph{optimal length} (hereafter just \emph{length}) $|x|$ of a position $x \in V$ is: the {minimum} number of moves necessary to force $x$'s win if $\lambda(x) = \WON$ (i.e., $x$'s value becomes known after its first follower is
  \LOST); the {maximum} number of moves that $x$'s losing can be delayed if $\lambda(x) = \LOST$ (i.e., $x$'s value
  is known after its last follower is \WON); and $\infty$
  (denoting infinite play) if $\lambda(x) = \DRAWN$.
\end{definition}
Optimal position length corresponds to the classic game-theoretic notion of
\emph{optimal play}: players try to win as quickly or lose as slowly
as possible while avoiding blunders. The length of a position can be
computed using AFP (or equivalently, backward induction) by simply using the iteration/state number in which the position's value becomes first known (starting at 0)
\cite{kohler_first-order_2013,BowersXiaLudaescherTAPP24}. 

\begin{definition}
  \label{def:edge-labels}
  Let $G = (V,E,\lambda)$ be a solved game. A \emph{provenance move labeling}
  $\Lambda : E \to \{\WON, \LOST, \DRAWN,
  \BLUNDER\}$ assigns labels for \emph{winning} (\WON),
  \emph{delaying} (\LOST), \emph{drawing} (\DRAWN), and
  \emph{blundering} (\BLUNDER) moves, such that:
  \begin{displaymath}
    \Lambda(x,y) \becomes
    \left\{
      \begin{array}{ll}
        \WON & \mbox{if } \lambda(x) = \WON \mbox{ and } \lambda(y) = \LOST \\
        \LOST & \mbox{if } \lambda(x) = \LOST \mbox{ and } \lambda(y) = \WON \\
        \DRAWN & \mbox{if } \lambda(x) = \DRAWN \mbox{ and } \lambda(y) = \DRAWN \\
        \BLUNDER & \mbox{otherwise} \\
      \end{array}
    \right .
  \end{displaymath}
  A \emph{solved game with move labels} is denoted
  $G = (V, E, \lambda, \Lambda)$.
\end{definition}

The above move types depend on the value $\lambda(x)$ of the move's origin  $x$ and the value  $\lambda(y)$ of its follower position $y$. If a position $x$ is won, 
there must be a \emph{winning} move to a follower $y$ that is lost for the opponent. 
Choosing any other follower (i.e., one that is drawn or won for the opponent) is a 
\emph{blunder}. On the other hand, if $x$ is drawn, there cannot be a lost follower (otherwise $x$ would be winning, not drawn). Instead one must find a drawn follower $y$ to keep the draw. Moving to a follower that is won (for the opponent) is another kind of \emph{blunder}. Finally, if $x$ is lost, there are no lost or drawn followers (otherwise, $x$ would not be lost) and the only option is a \emph{delaying} move to a position $y$ that is won for the opponent. 

Move labels (edge types) are used directly to define the \emph{actual provenance} $\Aprov(x)$ of a position $x$ in a solved game~\cite{BowersXiaLudaescherTAPP24}.

\begin{definition}
  Given a solved game $G = (V,E,\lambda, \Lambda)$ with move labels, the
  \emph{actual provenance} $\Aprov(x)$ of a position $x$ is the
  subgraph reachable from $x$ by only following \WON-,
  \LOST-, and \DRAWN-labeled moves. In particular, blundering
  (\BLUNDER) moves must be ignored.
\end{definition}

Winning moves can be further categorized as either \emph{primary} or
\emph{secondary} based on optimal play. A \emph{primary winning move}
(labeled \WONPR) from $x$ is a winning move that is part of a
shortest-length win for $x$. A \emph{secondary winning move} (labeled
\WONSC) from $x$ is a non-shortest winning move. Both the
AFP-based and backward-induction algorithms for solving games can be
instrumented to compute all primary (\WONPR) and
delaying (\LOST) edge labels \cite{BowersXiaLudaescherTAPP24}. The remaining edge
labels can be obtained using Definition~\ref{def:edge-labels}.

\begin{definition}
  Given a solved game $G = (V,E,\lambda,\Lambda)$ with move labeling $\Lambda$ (and labels for \WONPR and \WONSC), the \emph{primary provenance} $\Pprov(x)$ of a position $x$ is the
  subgraph reachable from $x$ by only following \WONPR-,
  \LOST-, and \DRAWN-labeled edges. Blundering
  (\BLUNDER) and secondary winning moves (\WONSC) are
  ignored.
\end{definition}

\subsection{The Skeptic's Argumentation Game}

The \emph{Skeptic's Argumentation Game} (SAG) is a direct interpretation of win-move for AFs under the grounded semantics. SAG uses the correspondence between $P_{AF1}$ and $P_{WM}$: attacks are traversed (as moves) in the ``attacked-by'' direction, accepted (\IN) arguments correspond to \LOST positions, defeated (\OUT) arguments correspond to \WON positions, and undecided (\UNDEC) arguments correspond to \DRAWN positions. Through this correspondence, SAG allows for the direct transfer of game provenance results to the AF setting. SAG is played as follows.

\mypara{A Skeptic's Perspective} Player~I, the \emph{Skeptic}, argues that
a node $x$ in AF is a \emph{defeated} argument. To this end, the
Skeptic claims that there \emph{exists} an attacker $y$ in the
AF-graph which itself is accepted and $y$ attacks $x$ (or equivalently
$x$ is \emph{attacked-by} $y$). This corresponds to Player~I choosing one
of the possible moves (from $x$) in the game.  Player~II, the
\emph{Optimist}, begs to differ and makes the counter claim that
\emph{all} attackers of $x$ are defeated, including $y$. Play continues until an argument is reached that is unattacked. 

\begin{figure}[t]
  \centering
   \includegraphics[width=.95\columnwidth]{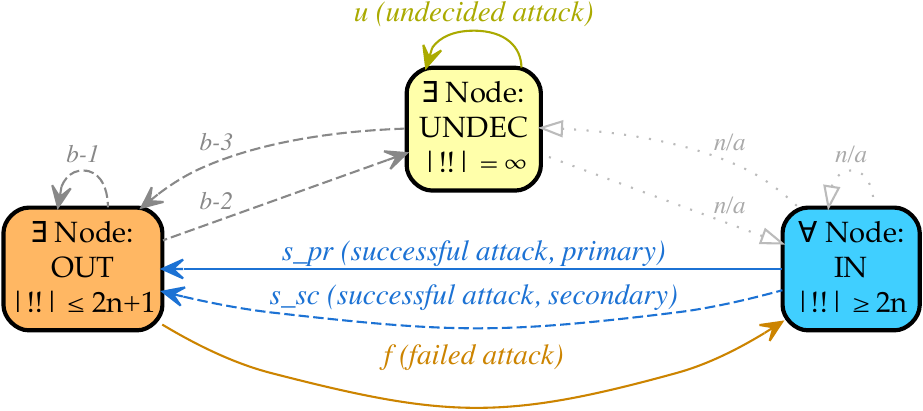}
  \caption{\small Classification of \emph{attack types}: Argument $x$ is \OUT if there \emph{exists} ($\exists$) a \emph{primary} or \emph{secondary} attack from an argument $y$ that is \IN ($y$'s attack is \emph{successful}). Argument $x$ is \IN if \emph{all} ($\forall$) of its attackers are \OUT (all  attacks on $x$ \emph{fail}).  The \emph{length} of $x$ indicates the number of attacks in an optimal dialogue (game): A defeat can be \emph{forced} from an
    \OUT argument in \emph{at most} $2n+1$ attacks; acceptance can be delayed for \emph{at least} $2n$ attacks. (``!!''  denotes a brilliant move in chess.)  If no side can force a win from $x$,
    then $x$ is \UNDEC: there \emph{exists} ($\exists$) an attack that prevents defeat (by repeating attacks and cyclic arguments). Attack types
    ``\emph{b-1}, \emph{b-2}, \emph{b-3}'' are blunders and thus irrelevant for determining argument values. Label ``{\emph{n/a}}'' indicates attack types that cannot
    exist.}
    \label{fig:attack_types}
\end{figure}

By employing the standard machinery of win-move, the definitions of games, solutions, the AFP-based (backward induction) algorithm, length, and actual and primary
provenance immediately apply to SAG. The \emph{type graph} in
Figure~\ref{fig:attack_types} summarizes the overall provenance structure
of AFs with respect to argument values and attack types. The seven
attack types are divided into provenance-\emph{relevant} attacks
(\emph{successful}, \emph{failed}, \emph{undecided}) and
provenance-\emph{irrelevant} attacks (three kinds of attack \emph{blunders}).  The \emph{successful} attacks are further
subdivided into \emph{primary} and \emph{secondary} attacks. \figref{fig:attack_types} also shows the three edge types that cannot exist in a grounded labeling. For example, an
argument would not be accepted if it were attacked by an accepted or
an undecided argument. Similarly, an undecided argument $x$ can never
be attacked by an accepted argument $y$, otherwise $x$ would be
defeated rather than undecided.

\section{Critical Attacks and Provenance Overlays}
\label{sect:critical}

This section extends provenance for grounded AFs to the stable semantics. Our approach relies on identifying minimal \emph{critical attack sets} that pinpoint the underlying assumptions made regarding argument acceptance within a stable solution. The critical attack sets highlight problematic attack edges that lead to undecided arguments within the original grounded extension. Each critical attack set is tied to a specific stable solution. For a stable solution, if the corresponding critical attacks are removed from the AF, producing a modified version, the modified AF's grounded solution has the same argument values as the stable solution of the original AF. Critical attack sets represent a new form of attack provenance, highlighting the choices and assumptions of the stable solution. When the critical attacks are suspended (ignored or removed), SAG's game-based provenance labelings can be applied to help explain the ``ambiguous regions'' of the stable solution.

\begin{definition} Let $G = (V,E)$, $S_0 \subseteq V$ be the grounded extension, and $S_i \subseteq V$ be a stable extension of $G$ (for $i \ge 1$). A \emph{critical attack set} $\Delta_{i,j} \subseteq E$ of $S_i$ is a minimal set of attacks such that $S_i$ is the grounded extension of $G' = (V, E \setminus \Delta_{i,j})$. The set $\mathbf{\Delta_i} = \{\Delta_{i,1}, \Delta_{i,2}, \dots, \Delta_{i,n}\}$ contains all such minimal critical attack sets of $S_i$.
\end{definition}

\begin{figure}[!t]
  \centering
  \subfloat[The original grounded solution $S_0$]{
    \includegraphics[width=.26\textwidth]{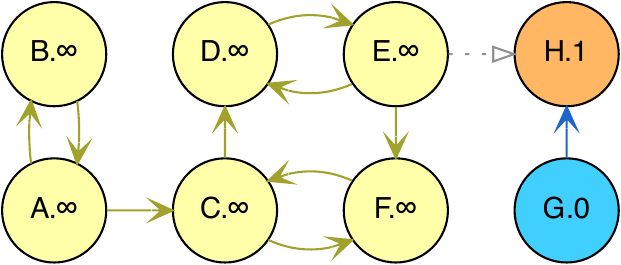}
    \label{fig:ex2-wf}
  }
  \\
  \subfloat[Stable solution $S_1$ with critical set $\Delta_{1,1}$]{
    \includegraphics[width=.255\textwidth]{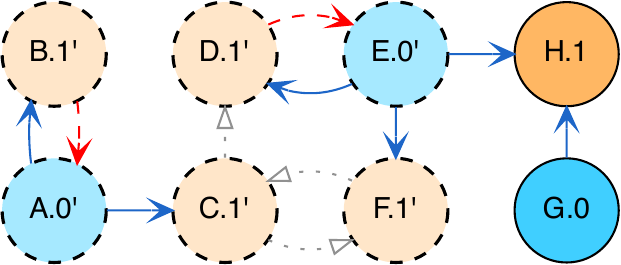}
    \label{fig:ex2-stb-1}
  }
  \\
   \subfloat[Stable solution $S_2$ with critical set $\Delta_{2,1}$]{
    \includegraphics[width=.255\textwidth]{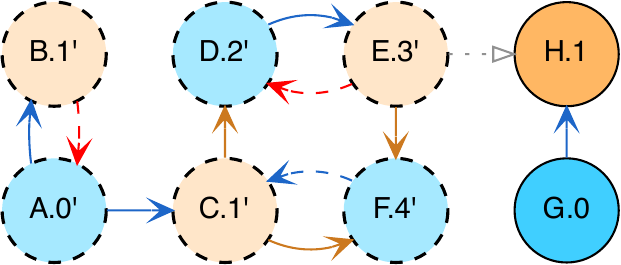}
    \label{fig:ex2-stb-2}
  }
  \\
  \subfloat[Stable solution $S_3$ with critical sets $\Delta_{3,1}$, $\Delta_{3,2}$]{
    \begin{tabular}{c}
    \includegraphics[width=.255\textwidth]{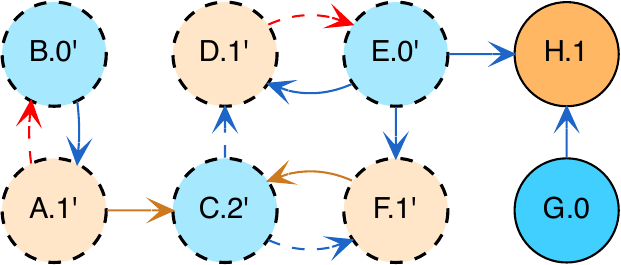} \\[5pt]
    \includegraphics[width=.255\textwidth]{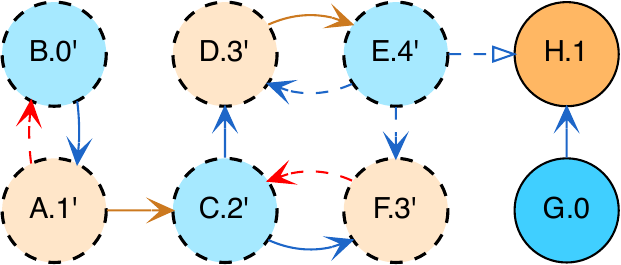}
    \end{tabular}
    \label{fig:ex2-stb-3}
  }
  \\
  \subfloat[Stable solution $S_4$ with critical sets $\Delta_{4,1}$, $\Delta_{4,2}$]{
    \begin{tabular}{c}
    \includegraphics[width=.255\textwidth]{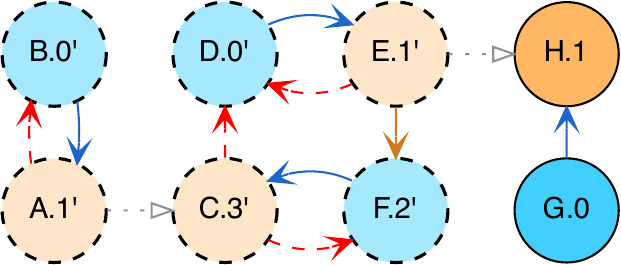} \\[5pt]
    \includegraphics[width=.255\textwidth]{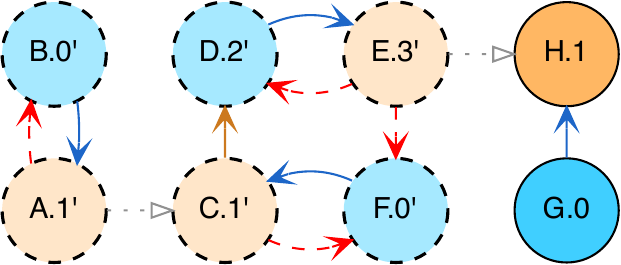}
    \end{tabular}
    \label{fig:ex2-stb-4}
  }
  \caption{\small Example AF with four stable solutions $S_1$--$S_4$: (a) grounded solution $S_0$; and (b-c) overlays for each stable solutions $S_1$--$S_4$.}
  \label{fig:ex2}
\end{figure}

\mypara{Provenance Overlays for Stable Solutions} A \emph{provenance overlay} consists of additional argument, length, and edge provenance labels for a given stable solution. Consider the example of \Figref{fig:ex2}. The grounded solution $S_0$ is shown in \Figref{fig:ex2-wf}, and its four stable solutions $S_1$--$S_4$ (as overlays) in \figref{fig:ex2-stb-1}--\ref{fig:ex2-stb-4}, respectively. An overlay consists of the original \IN and \OUT argument labels from the grounded solution $S_0$. In place of \UNDEC argument labels in $S_0$ (e.g., \pos{A}--\pos{F} in \Figref{fig:ex2-wf}), the overlay contains the argument labels \OIN and \OOUT for the corresponding accepted and defeated arguments in the stable solution $S_i$, shown with dashed lines and using pale blue and orange, respectively. Overlays also contain the grounded attack labels but with \UNDEC replaced by corresponding \OIN (both secondary and primary), \OOUT, and \OBLUNDER attack labels, obtained from the grounded attack labeling of the AF after removing the edges in the critical attack set. In the overlay, critical attacks are labeled as \textsf{\textsc{critical}} and displayed using dashed red edges. Length values in an overlay are taken from the grounded solution of the modified AF (with critical attack edges suspended). For example, in \Figref{fig:ex2-stb-1}, argument \pos D is displayed as \pos{D.1{\tt '}}, denoting that its length value in the overlay is 1. Arguments with the same length values in the original grounded solution are denoted as such, e.g., \pos{H.1} is the same in both the original and the modified grounded labeling.

\mypara{Critical Attack Sets and Overlay Examples} \Figref{fig:ex2-stb-1} and \ref{fig:ex2-stb-2} represent stable solutions $S_1$ and $S_2$, where argument \pos A was assumed accepted, as denoted by the critical attack edge $\pos B {\to} A$ in both figures. In $S_1$ (\Figref{fig:ex2-stb-1}) an additional choice was made to accept \pos E as shown by critical attack edge $\pos D {\to} \pos E$. Alternatively, in $S_2$ (\Figref{fig:ex2-stb-2}), the acceptance of \pos D is explained by the critical attack $\pos E {\to} \pos D$. Furthermore, in $S_2$ the choice of \pos D leads to \pos F being accepted (since \pos D defends \pos F), whereas in $S_1$, the two chosen arguments (\pos A and \pos E) do not lead to additional argument acceptance.

\Figref{fig:ex2-stb-3} and \ref{fig:ex2-stb-4} show overlays for stable solutions $S_3$ and $S_4$, respectively. $S_3$ and $S_4$ each represent the case where \pos B is assumed to be accepted. Both stable solutions have two distinct (minimal) critical attack sets. In $S_3$, both of its critical attack sets contain two edges, whereas $S_4$'s critical attack sets contain four edges. In $S_3$, the first critical attack set assumes \pos E (top of \Figref{fig:ex2-stb-3}) and the second critical attack set assumes \pos C (bottom of \Figref{fig:ex2-stb-3}). Note that assuming both \pos E and \pos C is unnecessary since, e.g., \pos E defends \pos C, and vice versa. In $S_4$, \pos D and \pos F are assumed in both critical attack sets, however, the two sets differ with respect to the specific attacks that can be suspended. In this case, \pos D alone does not imply \pos F, and vice versa.  As shown, the distinct critical attack sets for both $S_3$ and $S_4$ result in significantly different argument lengths in the overlay.  

\begin{figure}[!t]
  \centering
  \subfloat[Layered display of $S_1$ with critical set $\Delta_{1,1}$]{
    \includegraphics[scale=.38]{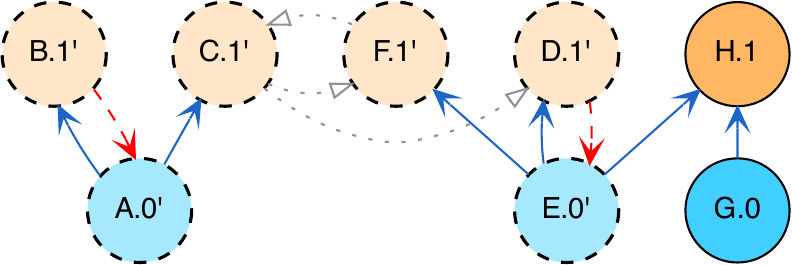}
    \label{fig:ex2-stb-layered-1}
  }
  \\
  \subfloat[Layered display of $S_2$ with critical set $\Delta_{2,1}$]{
    \hspace{0.5in}\includegraphics[scale=.38]{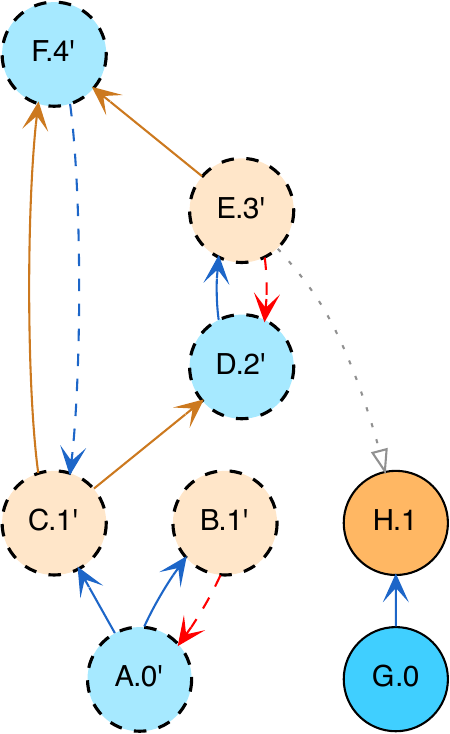}\hspace{0.5in}
    \label{fig:ex2-stb-layered-2}
  }  
  \\
   \subfloat[Layered displays of $S_3$ with critical sets $\Delta_{3,1}$ (left) and $\Delta_{3,2}$ (right)]{
    \includegraphics[scale=.38]{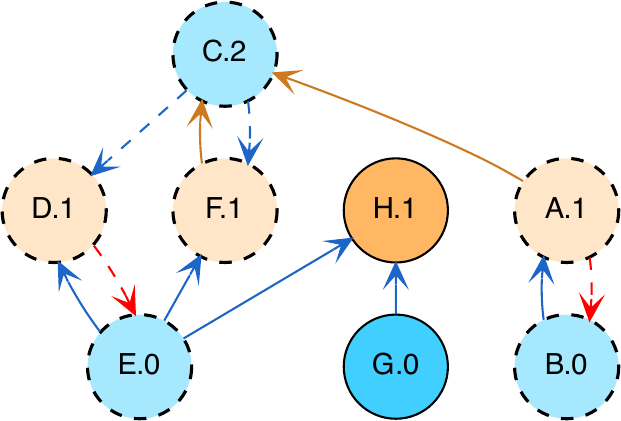} 
    \hspace{2em}
    \includegraphics[scale=.35]{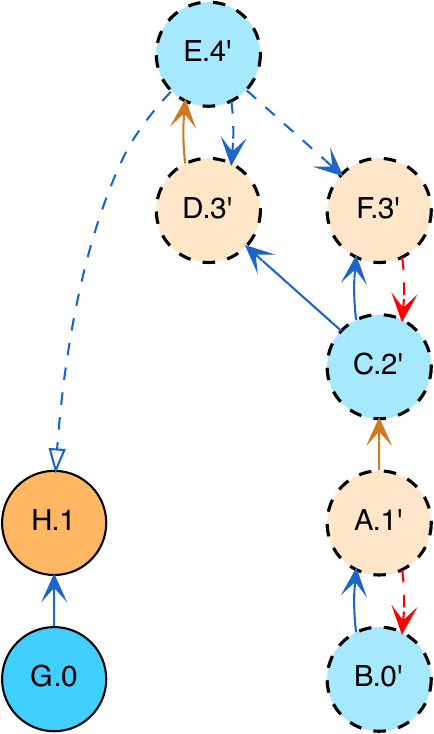}
    \label{fig:ex2-stb-layered-3}
  }
  \\
  \subfloat[Layered displays of $S_4$ with critical sets $\Delta_{4,1}$ (left) and $\Delta_{4,2}$ (right)]{
    \includegraphics[scale=.38]{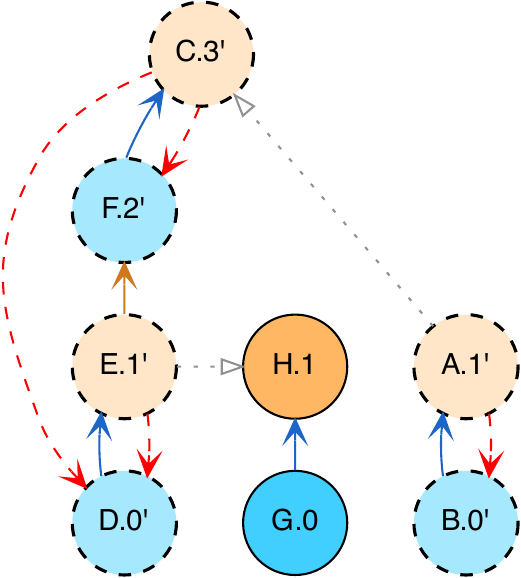}
    \hspace{2em}
    \includegraphics[scale=.38]{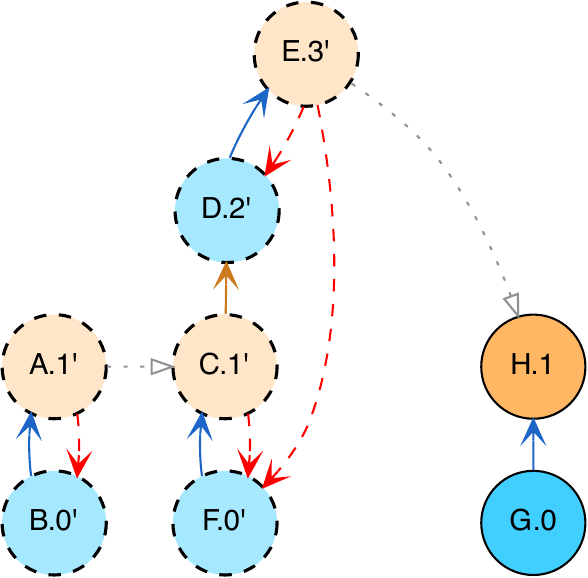}
    \label{fig:ex2-stb-layered-4}
  }
 \caption{\small Layered visualizations of the $S_1$--$S_4$ overlays from \Figref{fig:ex2}.
  }
 \label{fig:ex2-layered}
\end{figure}

\mypara{Layered Renderings of Overlays} \Figref{fig:ex2-layered} gives the layered visualizations of the stable solution overlays of \Figref{fig:ex2}. As shown, the provenance structure varies significantly for each stable solution. The same is true for the different critical attack sets of $S_3$ (especially) and $S_4$. For example, argument \pos E in the first critical attack set of $S_3$ (\Figref{fig:ex2-stb-layered-3}, left) has no attackers (and is thus in the lowest layer of the graph), whereas in the second attack set (\Figref{fig:ex2-stb-layered-3}, right), \pos E is in the highest layer of the graph such that the provenance of \pos E contains \pos D (as its only immediate attacker), \pos C (its immediate defender), \pos A (as \pos C's only attacker), and finally \pos B (a defender of \pos C). The layered visualizations of the different critical attack sets of a given stable solution can provide further insight into the stable solution itself and, e.g., in the case of diagnosis, provide additional information for selecting an appropriate repair.   

\mypara{Computing Critical Attack Sets} For a given AF graph $G = (V,E)$, a stable extension $S_i$, and $G$'s grounded solution $S_0$, the critical attack sets of $S_i$ can be computed naively as follows:   
\begin{enumerate}
    \item Find each candidate subset $D_{i,j} \subseteq E$ of \UNDEC attacks $X{\to}Y$ in $S_0$ where the grounded extension of $G' = (V, E \setminus D_{i,j})$ is equivalent to the stable extension $S_i$; and
    \item Select the minimal candidates $D_{i,1}, \dots, D_{i,k}$ to be the critical attack sets $\Delta_{i,1}, \dots, \Delta_{i,k}$ of $S_i$.
\end{enumerate}
The problem of computing the grounded extension is PTIME, however, in the worst case, all subsets of the edges in $G$ may need to be checked in step (1). An implementation of the above algorithm is shown in \Figref{fig:code}. The program is written using the Answer Set Programming (ASP) syntax of Clingo\footnote{\href{https://potassco.org/clingo}{https://potassco.org/clingo}}. The program finds all cardinality minimal critical attack sets of an AF. Alternatively, to find the subset-minimal critical attack sets, the program can be easily modified to find all (potentially non-minimal) critical attack sets, and then employ a post-processing step to select only those critical attack sets that are subset minimal. 

The implementation of \Figref{fig:code} expects (as a separate input file) a set of $\texttt{attacks}(X,Y)$ facts representing an input AF graph as well as the desired stable solution. Note that the stable solutions can also easily be computed with Clingo using the $P_{AF2}$ rules, e.g., as an initial step. The implementation proceeds as follows. Lines 2--3 extract the arguments from the input facts. Lines 6--8 compute the grounded labeling for $S_0$. The rule on line 6 uses a conditional literal expression $\texttt{out0}(Y) : \texttt{attacks}(Y,X)$ that is interpreted as $(\forall Y)(\texttt{attacks}(Y,X) \to \texttt{out0}(Y))$. Line 11 is a  choice rule, which ``guesses'' \UNDEC edges to include in the critical attack set. Line 14 then ignores the chosen critical attacks in a new $\texttt{attacks1}$ relation. Lines 17--19 compute the grounded labeling of the modified graph with the critical attack edges removed. Line 22 is a constraint requiring that no \UNDEC nodes exist in the new grounded solution to ensure that a partial stable model is not selected. Line 25 uses an aggregate function to count the number of critical edges selected in line 11. Finally, line 26 only selects answer sets that are cardinality minimal by using Clingo's {\tt minimize} directive.


\lstdefinelanguage{CLINGO}
{
  belowcaptionskip=1\baselineskip,
  breaklines=true,
  frame=topbottom,
  numbers=left, 
  basicstyle=\footnotesize\ttfamily,
  keywordstyle=\color{black},
  commentstyle=\itshape\color{gray},
  identifierstyle=\color{black},
  stringstyle=\color{red!50!black},
  showstringspaces=false,
  sensitive=false,  
  keywords={not},
  morecomment=[l]{\%},
  morestring=[b]",
  emphstyle=\color{orange}  
}

\begin{figure}
\begin{lstlisting}[language=CLINGO]
% Argument nodes in the graph
arg(X) :- attacks(X,_).
arg(X) :- attacks(_,X).

% Compute the initial grounded extension 
in0(X) :- arg(X), out0(Y) : attacks(Y,X).
out0(X) :- attacks(Y,X), in0(Y).
undec0(X) :- arg(X), not in0(X), not out0(X).

% Guess critical attacks: subset of UNDEC attacks
{critical(Y,X)} :- attacks(Y,X), undec0(Y), undec0(X). 

% Keep attacks if they are not critical 
attacks1(Y,X) :- attacks(Y,X), not critical(Y,X).

% Compute grounded labeling without critical attacks 
in(X) :- arg(X), out(Y) : attacks1(Y,X).
out(X) :- attacks1(Y,X), in(Y).
undec(X) :- arg(X), not in(X), not out(X).

% Eliminate candidate solutions having UNDEC nodes
:- undec(_).

% Select solutions with fewest critical attacks
critical_cnt(N) :- N = #count{(X,Y) : critical(X,Y)}.
#minimize {N : critical_cnt(N)}.
\end{lstlisting}
\caption{Example Clingo (ASP) program for computing the cardinality-minimal critical attack sets of a given AF (represented by an input {\tt attack} relation) and stable solution (expressed in the input as a set of ASP constraints).}
\label{fig:code}
\end{figure}


\section{Discussion}
\label{sect:discussion}

We further developed our prior work \cite{BowersXiaLudaescherTAPP24,BowersXiaLudaescherSAFA24} on provenance for AFs under the grounded semantics by extending it to AF solutions under the stable semantics. Our work is based on the observation that certain critical attacks form the basis of choices (or assumptions) made in a stable solution, and can be used as part of an explanation. By suspending (i.e., temporarily removing) these critical attacks, the natural provenance of acceptance and defeat used in grounded solutions can be  applied to stable solutions. We  used this approach to develop a new notion of \emph{solution overlay}, which provides a layered and provenance-annotated visualization of stable solutions. An overlay can be used to  inform users of the underlying choices behind different stable solutions. The critical attacks also suggest minimal repairs for users who wish to resolve unexpected ambiguity, e.g., when they expect a single 2-valued grounded solution.

Critical attack sets are closely related to the AF explanation approach of Baumann and Ulbricht \cite{BaumannUlbricht2021} for the complete semantics. Their approach similarly utilizes grounded extensions and considers an explanation of a complete (including stable) extension as three pairwise disjoint sets: (1) the accepted arguments of the grounded extension; (2) the set of arguments on even cycles that represent choices; and (3) the remaining accepted arguments that result from applying the choices. 

Our sets of critical attack edges provide a compatible but more nuanced form of explanation, e.g., as illustrated by the $S_3$ and $S_4$ solutions in \Figref{fig:ex2}: Here, not only are the chosen arguments identified (as the targets of the suspended critical attacks), but the attacks themselves are provided for further analysis and repair. The critical attacks are also necessary for the resulting layered display of an overlay. The PyArg \cite{odekerken2023pyarg} tool for analyzing and visualizing AFs  provides extension-based explanations for acceptance under different AF semantics. These explanations are based on the framework of Borg and Bex \cite{BorgBex2024}, which uses direct and indirect defense to construct an explanation as a single set of arguments. Both skeptical and credulous explanations are considered (as the intersection and union of defense sets, respectively, across extensions for a particular accepted argument). In both cases, the notion of choice is not part of the explanations. 

Our work is also similar to enforcement in AFs \cite{Baumann_etal_2021}, where AF modifications (typically in the form of additional arguments and attacks) are considered to obtain a desired argument labeling. 
 
In future work, we plan to extend the notion of critical attack sets to \emph{complete} (and thus also \emph{preferred}) semantics, and develop extensions of the PyArg framework to identify critical attack sets and compute overlays. To support use cases like the one described in \Figref{fig:3panel}, \textsc{AF-Xray}, a prototype based on PyArg,  has been implemented and applied to legal reasoning \cite{xia2025afxray}. We are also interested in combining causality-based approaches for explanation in AFs~\cite{Gianvincenzo_etal_2024} with the AF provenance and visualization approaches described in this paper. Finally, we plan to explore the application of the approaches presented here to the work on instantiation of AFs from logic programs \cite{dung1995acceptability,caminada2015equivalence}, where AF repair can be translated to repairs in the underlying (non-monotonic and/or defeasible) logic program represented by the AF.

\begin{acks}
This work supported in part by the Joint Research and Innovation Seed Grants Program between the University of Illinois System and Cardiff University under the seed grant ``\emph{XAI-CA: Explainable AI via Computational Argumentation.}''
\end{acks}

\bibliographystyle{ACM-Reference-Format}
\bibliography{tapp25}


\begin{thebibliography}{26}


\ifx \showCODEN    \undefined \def \showCODEN     #1{\unskip}     \fi
\ifx \showDOI      \undefined \def \showDOI       #1{#1}\fi
\ifx \showISBNx    \undefined \def \showISBNx     #1{\unskip}     \fi
\ifx \showISBNxiii \undefined \def \showISBNxiii  #1{\unskip}     \fi
\ifx \showISSN     \undefined \def \showISSN      #1{\unskip}     \fi
\ifx \showLCCN     \undefined \def \showLCCN      #1{\unskip}     \fi
\ifx \shownote     \undefined \def \shownote      #1{#1}          \fi
\ifx \showarticletitle \undefined \def \showarticletitle #1{#1}   \fi
\ifx \showURL      \undefined \def \showURL       {\relax}        \fi
\providecommand\bibfield[2]{#2}
\providecommand\bibinfo[2]{#2}
\providecommand\natexlab[1]{#1}
\providecommand\showeprint[2][]{arXiv:#2}

\bibitem[Alfano et~al\mbox{.}(2024)]%
        {Gianvincenzo_etal_2024}
\bibfield{author}{\bibinfo{person}{G. Alfano}, \bibinfo{person}{S. Greco}, \bibinfo{person}{F. Parisi}, {and} \bibinfo{person}{I. Trubitsyna}.} \bibinfo{year}{2024}\natexlab{}.
\newblock \showarticletitle{\href{https://doi.org/10.24963/kr.2024/2}{Counterfactual and semifactual explanations in abstract argumentation: formal foundations, complexity and computation}}. In \bibinfo{booktitle}{\emph{KR}}.
\newblock


\bibitem[Baroni et~al\mbox{.}(2018)]%
        {baroni_handbook_2018}
\bibfield{author}{\bibinfo{person}{P. Baroni}, \bibinfo{person}{D. Gabbay}, \bibinfo{person}{M. Giacomin}, {and} \bibinfo{person}{L. van~der Torre}.} \bibinfo{year}{2018}\natexlab{}.
\newblock \bibinfo{booktitle}{\emph{\href{https://philpapers.org/rec/BARHOF}{Handbook of {Formal} {Argumentation}}}}.
\newblock \bibinfo{publisher}{London: College Publications}.
\newblock


\bibitem[Baumann et~al\mbox{.}(2021)]%
        {Baumann_etal_2021}
\bibfield{author}{\bibinfo{person}{R. Baumann}, \bibinfo{person}{S. Doutre}, \bibinfo{person}{J.-G. Mailly}, {and} \bibinfo{person}{J.P. Wallner}.} \bibinfo{year}{2021}\natexlab{}.
\newblock \showarticletitle{\href{https://hal.science/hal-03541704v1}{Enforcement in Formal Argumentation}}.
\newblock In \bibinfo{booktitle}{\emph{Handbook of Formal Argumentation}}. Vol.~\bibinfo{volume}{2}. \bibinfo{pages}{445--510}.
\newblock


\bibitem[Baumann and Ulbricht(2021)]%
        {BaumannUlbricht2021}
\bibfield{author}{\bibinfo{person}{R. Baumann} {and} \bibinfo{person}{M. Ulbricht}.} \bibinfo{year}{2021}\natexlab{}.
\newblock \showarticletitle{\href{https://doi.org/10.24963/kr.2021/11}{Choices and their Consequences - Explaining Acceptable Sets in Abstract Argumentation Frameworks}}. In \bibinfo{booktitle}{\emph{{KR}}}. \bibinfo{pages}{110--119}.
\newblock


\bibitem[Bench-Capon(2002)]%
        {bench-capon_representation_2002}
\bibfield{author}{\bibinfo{person}{T. Bench-Capon}.} \bibinfo{year}{2002}\natexlab{}.
\newblock \showarticletitle{\href{https://jurix.nl/pdf/j02-11.pdf}{Representation of {Case} {Law} as an {Argumentation} {Framework}}}. In \bibinfo{booktitle}{\emph{JURIX}}. \bibinfo{pages}{103--112}.
\newblock


\bibitem[Bench-Capon(2020)]%
        {BenchCapon20}
\bibfield{author}{\bibinfo{person}{T. Bench-Capon}.} \bibinfo{year}{2020}\natexlab{}.
\newblock \showarticletitle{\href{https://doi.org/10.3233/AAC-190477}{Before and after Dung: Argumentation in {AI} and Law}}.
\newblock \bibinfo{journal}{\emph{Argument \& Computation.}} \bibinfo{volume}{11}, \bibinfo{number}{1-2} (\bibinfo{year}{2020}), \bibinfo{pages}{221--238}.
\newblock


\bibitem[Bench-Capon and Modgil(2009)]%
        {BenchCaponModgil09}
\bibfield{author}{\bibinfo{person}{T. Bench-Capon} {and} \bibinfo{person}{S. Modgil}.} \bibinfo{year}{2009}\natexlab{}.
\newblock \showarticletitle{\href{https://doi.org/10.1145/1568234.1568248}{Case law in extended argumentation frameworks}}. In \bibinfo{booktitle}{\emph{ICAIL}}. \bibinfo{pages}{118–127}.
\newblock


\bibitem[Borg and Bex(2024)]%
        {BorgBex2024}
\bibfield{author}{\bibinfo{person}{A. Borg} {and} \bibinfo{person}{F. Bex}.} \bibinfo{year}{2024}\natexlab{}.
\newblock \showarticletitle{\href{https://doi.org/10.1016/j.ijar.2024.109143}{Minimality, necessity and sufficiency for argumentation and explanation}}.
\newblock \bibinfo{journal}{\emph{Int. J. Approx. Reasoning}} \bibinfo{volume}{168}, \bibinfo{number}{C} (\bibinfo{date}{May} \bibinfo{year}{2024}).
\newblock


\bibitem[Bowers et~al\mbox{.}(2024a)]%
        {BowersXiaLudaescherSAFA24}
\bibfield{author}{\bibinfo{person}{S. Bowers}, \bibinfo{person}{Y. Xia}, {and} \bibinfo{person}{B. Lud{\"{a}}scher}.} \bibinfo{year}{2024}\natexlab{a}.
\newblock \showarticletitle{\href{https://ceur-ws.org/Vol-3757/paper8.pdf}{The Skeptic's Argumentation Game or: Well-Founded Explanations for Mere Mortals}}. In \bibinfo{booktitle}{\emph{Workshop on Systems and Algorithms for Formal Argumentation (SAFA)}} \emph{(\bibinfo{series}{{CEUR}}, Vol.~\bibinfo{volume}{3757})}. \bibinfo{pages}{104--118}.
\newblock


\bibitem[Bowers et~al\mbox{.}(2024b)]%
        {BowersXiaLudaescherTAPP24}
\bibfield{author}{\bibinfo{person}{S. Bowers}, \bibinfo{person}{Y. Xia}, {and} \bibinfo{person}{B. Lud{\"{a}}scher}.} \bibinfo{year}{2024}\natexlab{b}.
\newblock \showarticletitle{\href{https://doi.org/10.1109/EuroSPW61312.2024.00073}{On the Structure of Game Provenance and its Applications}}. In \bibinfo{booktitle}{\emph{TaPP}}. \bibinfo{pages}{602--609}.
\newblock


\bibitem[Cabrio and Villata(2012)]%
        {Cabrio2012}
\bibfield{author}{\bibinfo{person}{E. Cabrio} {and} \bibinfo{person}{S. Villata}.} \bibinfo{year}{2012}\natexlab{}.
\newblock \showarticletitle{\href{https://aclanthology.org/P12-2041/}{Combining Textual Entailment and Argumentation Theory for Supporting Online Debates Interactions}}. In \bibinfo{booktitle}{\emph{Annual Meeting of the Association for Computational Linguistics}}. \bibinfo{pages}{208--212}.
\newblock


\bibitem[Caminada(2006)]%
        {caminada2006issue}
\bibfield{author}{\bibinfo{person}{M. Caminada}.} \bibinfo{year}{2006}\natexlab{}.
\newblock \showarticletitle{\href{https://doi.org/10.1007/11853886_11}{On the {{Issue}} of {{Reinstatement}} in {{Argumentation}}}}. In \bibinfo{booktitle}{\emph{Logics in {{Artificial Intelligence}}}}. \bibinfo{pages}{111--123}.
\newblock


\bibitem[Caminada(2018)]%
        {caminada2018argumentation_as_discussion}
\bibfield{author}{\bibinfo{person}{M. Caminada}.} \bibinfo{year}{2018}\natexlab{}.
\newblock \showarticletitle{{Argumentation Semantics as Formal Discussion}}.
\newblock See \citeN{baroni_handbook_2018}, Chapter~10, \bibinfo{pages}{487--518}.
\newblock


\bibitem[Caminada and Dunne(2020)]%
        {caminada_strong_2020}
\bibfield{author}{\bibinfo{person}{M. Caminada} {and} \bibinfo{person}{P. Dunne}.} \bibinfo{year}{2020}\natexlab{}.
\newblock \showarticletitle{\href{https://content.iospress.com/articles/argument-and-computation/aac190463}{{Strong Admissibility Revisited: Theory and Applications}}}.
\newblock \bibinfo{journal}{\emph{Argument \& Computation}} \bibinfo{volume}{10}, \bibinfo{number}{3} (\bibinfo{year}{2020}), \bibinfo{pages}{277--300}.
\newblock


\bibitem[Caminada et~al\mbox{.}(2015)]%
        {caminada2015equivalence}
\bibfield{author}{\bibinfo{person}{M. Caminada}, \bibinfo{person}{S. S{\'a}}, \bibinfo{person}{J. Alc{\^a}ntara}, {and} \bibinfo{person}{W. Dvo{\v r}{\'a}k}.} \bibinfo{year}{2015}\natexlab{}.
\newblock \showarticletitle{\href{https://doi.org/10.1016/j.ijar.2014.12.004}{On the Equivalence between Logic Programming Semantics and Argumentation Semantics}}.
\newblock \bibinfo{journal}{\emph{Approx.\ Reasoning}}  \bibinfo{volume}{58} (\bibinfo{year}{2015}), \bibinfo{pages}{87--111}.
\newblock


\bibitem[Dung(1995)]%
        {dung1995acceptability}
\bibfield{author}{\bibinfo{person}{P.M. Dung}.} \bibinfo{year}{1995}\natexlab{}.
\newblock \showarticletitle{\href{https://www.sciencedirect.com/science/article/pii/000437029400041X}{On the Acceptability of Arguments and Its Fundamental Role in Nonmonotonic Reasoning, Logic Programming and n-Person Games}}.
\newblock \bibinfo{journal}{\emph{AI}} \bibinfo{volume}{77}, \bibinfo{number}{2} (\bibinfo{year}{1995}), \bibinfo{pages}{321--357}.
\newblock


\bibitem[Gelfond and Lifschitz(1988)]%
        {gelfond_stable_1988}
\bibfield{author}{\bibinfo{person}{M. Gelfond} {and} \bibinfo{person}{V. Lifschitz}.} \bibinfo{year}{1988}\natexlab{}.
\newblock \showarticletitle{\href{http://www.cs.utexas.edu/users/ai-lab?gel88}{The Stable Model Semantics for Logic Programming}}. In \bibinfo{booktitle}{\emph{ILPS}}. \bibinfo{pages}{1070--1080}.
\newblock


\bibitem[K\"ohler et~al\mbox{.}(2013)]%
        {kohler_first-order_2013}
\bibfield{author}{\bibinfo{person}{S. K\"ohler}, \bibinfo{person}{B. Lud\"ascher}, {and} \bibinfo{person}{D. Zinn}.} \bibinfo{year}{2013}\natexlab{}.
\newblock \showarticletitle{\href{https://doi.org/10.1007/978-3-642-41660-6_20}{{First-Order Provenance Games}}}. In \bibinfo{booktitle}{\emph{In Search of Elegance in the Theory and Practice of Computation}}. \bibinfo{pages}{382--399}.
\newblock


\bibitem[Longo and Hederman(2013)]%
        {Longo2013}
\bibfield{author}{\bibinfo{person}{L. Longo} {and} \bibinfo{person}{L. Hederman}.} \bibinfo{year}{2013}\natexlab{}.
\newblock \showarticletitle{\href{https://doi.org/10.1007/978-3-319-02753-1_17}{Argumentation Theory for Decision Support in Health-Care: A Comparison with Machine Learning}}. In \bibinfo{booktitle}{\emph{Brain and Health Informatics}}. \bibinfo{pages}{168--180}.
\newblock


\bibitem[Lud{\"{a}}scher et~al\mbox{.}(2023)]%
        {ludascher2023games}
\bibfield{author}{\bibinfo{person}{B. Lud{\"{a}}scher}, \bibinfo{person}{S. Bowers}, {and} \bibinfo{person}{Y. Xia}.} \bibinfo{year}{2023}\natexlab{}.
\newblock \showarticletitle{\href{https://ceur-ws.org/Vol-3546/paper06.pdf}{{Games, Queries, and Argumentation Frameworks: Towards a Family Reunion}}}. In \bibinfo{booktitle}{\emph{Advances in Argumentation in AI}}, Vol.~\bibinfo{volume}{3546}. \bibinfo{publisher}{CEUR}.
\newblock


\bibitem[Odekerken et~al\mbox{.}(2023)]%
        {odekerken2023pyarg}
\bibfield{author}{\bibinfo{person}{D. Odekerken}, \bibinfo{person}{A.M. Borg}, {and} \bibinfo{person}{M. Berthold}.} \bibinfo{year}{2023}\natexlab{}.
\newblock \showarticletitle{\href{https://ceur-ws.org/Vol-3546/paper14.pdf}{{Demonstrating PyArg 2.0}}}. In \bibinfo{booktitle}{\emph{Advances in Argumentation in AI (AI$^3$)}}, Vol.~\bibinfo{volume}{3546}. \bibinfo{publisher}{CEUR}.
\newblock


\bibitem[Rahwan and Larson(2009)]%
        {Rahwan2009}
\bibfield{author}{\bibinfo{person}{I. Rahwan} {and} \bibinfo{person}{K. Larson}.} \bibinfo{year}{2009}\natexlab{}.
\newblock \bibinfo{booktitle}{\emph{\href{https://doi.org/10.1007/978-0-387-98197-0_16}{Argumentation and Game Theory}}}.
\newblock \bibinfo{pages}{321--339}.
\newblock


\bibitem[Van~Gelder(1993)]%
        {van1993alternating}
\bibfield{author}{\bibinfo{person}{A. Van~Gelder}.} \bibinfo{year}{1993}\natexlab{}.
\newblock \showarticletitle{\href{https://doi.org/10.1016/0022-0000(93)90024-Q}{{The Alternating Fixpoint of Logic Programs with Negation}}}.
\newblock \bibinfo{journal}{\emph{J. Comput. System Sci.}} \bibinfo{volume}{47}, \bibinfo{number}{1} (\bibinfo{year}{1993}), \bibinfo{pages}{185--221}.
\newblock


\bibitem[Van~Gelder et~al\mbox{.}(1991)]%
        {van1991well}
\bibfield{author}{\bibinfo{person}{A. Van~Gelder}, \bibinfo{person}{K.A. Ross}, {and} \bibinfo{person}{J.S. Schlipf}.} \bibinfo{year}{1991}\natexlab{}.
\newblock \showarticletitle{\href{http://doi.acm.org/10.1145/116825.116838}{The Well-founded Semantics for General Logic Programs}}.
\newblock \bibinfo{journal}{\emph{J. ACM}} \bibinfo{volume}{38}, \bibinfo{number}{3} (\bibinfo{year}{1991}), \bibinfo{pages}{619--649}.
\newblock


\bibitem[Xia et~al\mbox{.}(2024)]%
        {xia_layered_vis_demo_2024}
\bibfield{author}{\bibinfo{person}{Y. Xia}, \bibinfo{person}{D. Odekerken}, \bibinfo{person}{S. Bowers}, {and} \bibinfo{person}{B. Lud{\"{a}}scher}.} \bibinfo{year}{2024}\natexlab{}.
\newblock \showarticletitle{\href{{https://doi.org/10.3233/FAIA240346}}{Layered Visualization of Argumentation Frameworks}}. In \bibinfo{booktitle}{\emph{COMMA}}, Vol.~\bibinfo{volume}{388}. \bibinfo{pages}{373--374}.
\newblock


\bibitem[Xia et~al\mbox{.}(2025)]%
        {xia2025afxray}
\bibfield{author}{\bibinfo{person}{Y. Xia}, \bibinfo{person}{H. Zheng}, \bibinfo{person}{S. Bowers}, {and} \bibinfo{person}{B. Lud{\"a}scher}.} \bibinfo{year}{2025}\natexlab{}.
\newblock \showarticletitle{{{AF-\textsc{Xray}}: Visual Explanation and Resolution of Ambiguity in Legal Argumentation Frameworks}}. In \bibinfo{booktitle}{\emph{20th Intl.\ Conf.\ on Artificial Intelligence and Law (ICAIL)}}.
\newblock
\newblock
\shownote{\url{https://github.com/idaks/xray}}.


\end{thebibliography}
    
\end{document}